\pdfoutput=1

\documentclass[11pt]{article}

\usepackage[]{EMNLP2023}

\usepackage{times}
\usepackage{latexsym}

\usepackage[T1]{fontenc}

\usepackage[utf8]{inputenc}

\usepackage{microtype}

\usepackage{inconsolata}

\usepackage{bbm}
\usepackage{pifont}
\usepackage{booktabs}
\usepackage{makecell}
\usepackage{rotating}

\usepackage{subfig}
\usepackage{multirow}
\usepackage{arydshln}

\usepackage{graphicx}

\usepackage{algorithmicx}
\usepackage{algorithm}
\usepackage{algpseudocode}
\usepackage{amsmath} 
\usepackage{amssymb} 
\usepackage{mathtools}

\makeatletter
\newcommand{\removelatexerror}{\let\@latex@error\@gobble}
\makeatother

\algnewcommand{\LeftComment}[1]{\State \(\triangleright\) #1}

\usepackage{xcolor}

\title{DialCoT Meets PPO: Decomposing and Exploring Reasoning Paths in Smaller Language Models}

\author{
Chengcheng Han\textsuperscript{$\diamondsuit$}\thanks{~~Work done during an internship at Xiaobing.AI.}\quad \quad 
Xiaowei Du\textsuperscript{$\spadesuit$}\quad \quad 
Che Zhang\textsuperscript{$\heartsuit$}\\
\bf{
Yixin Lian\textsuperscript{$\spadesuit$}\quad
Xiang Li\textsuperscript{$\diamondsuit$}\quad
Ming Gao\textsuperscript{$\diamondsuit$}\textsuperscript{$\clubsuit$}\thanks{~~Corresponding author.}\quad
Baoyuan Wang\textsuperscript{$\spadesuit$}\footnotemark[2]
}\\
\textsuperscript{$\diamondsuit$}School of Data Science and Engineering, East China Normal University\\
\textsuperscript{$\spadesuit$}Xiaobing.AI \\
\textsuperscript{$\heartsuit$}School of Software \& Microelectronics, Peking University\\
\textsuperscript{$\clubsuit$}KLATASDS-MOE in School of Statistics, East China Normal University\\
\texttt{chengchenghan@stu.ecnu.edu.cn}\\
\texttt{\{duxiaowei,lianyixin,wangbaoyuan\}@xiaobing.ai}\\
\texttt{mmt@stu.pku.edu.cn} \\
\texttt{\{xiangli,mgao\}@dase.ecnu.edu.cn}
}

\begin{document}
\maketitle
\begin{abstract}
Chain-of-Thought~(CoT) prompting
has proven to be effective 
in enhancing the reasoning capabilities 
of Large Language Models~(LLMs) 
with at least 100 billion parameters. 
However, 
it is ineffective or even detrimental 
when applied to reasoning tasks 
in Smaller Language Models~(SLMs) 
with less than 10 billion parameters. 
To address this limitation, 
we introduce Dialogue-guided Chain-of-Thought~(DialCoT) 
which employs a dialogue format 
to generate intermediate reasoning steps, 
guiding the model toward the final answer. 
Additionally, 
we optimize the model's reasoning path selection 
using the Proximal Policy Optimization (PPO) algorithm, 
further enhancing its reasoning capabilities. 
Our method offers several advantages 
compared to previous approaches. 
Firstly, 
we transform the process of solving complex reasoning questions 
by breaking them down 
into a series of simpler sub-questions, 
significantly reducing the task difficulty 
and making it more suitable for SLMs.
Secondly,
we optimize the model's reasoning path selection 
through the PPO algorithm. 
We conduct comprehensive experiments 
on four arithmetic reasoning datasets, 
demonstrating that 
our method achieves significant performance improvements 
compared to state-of-the-art competitors. 
\footnote{Our code and dataset are publicly available at
\url{https://github.com/hccngu/DialCoT}.}
\end{abstract}

\section{Introduction}

\begin{figure}[!t]
    \centering
    \includegraphics[width=0.48\textwidth]{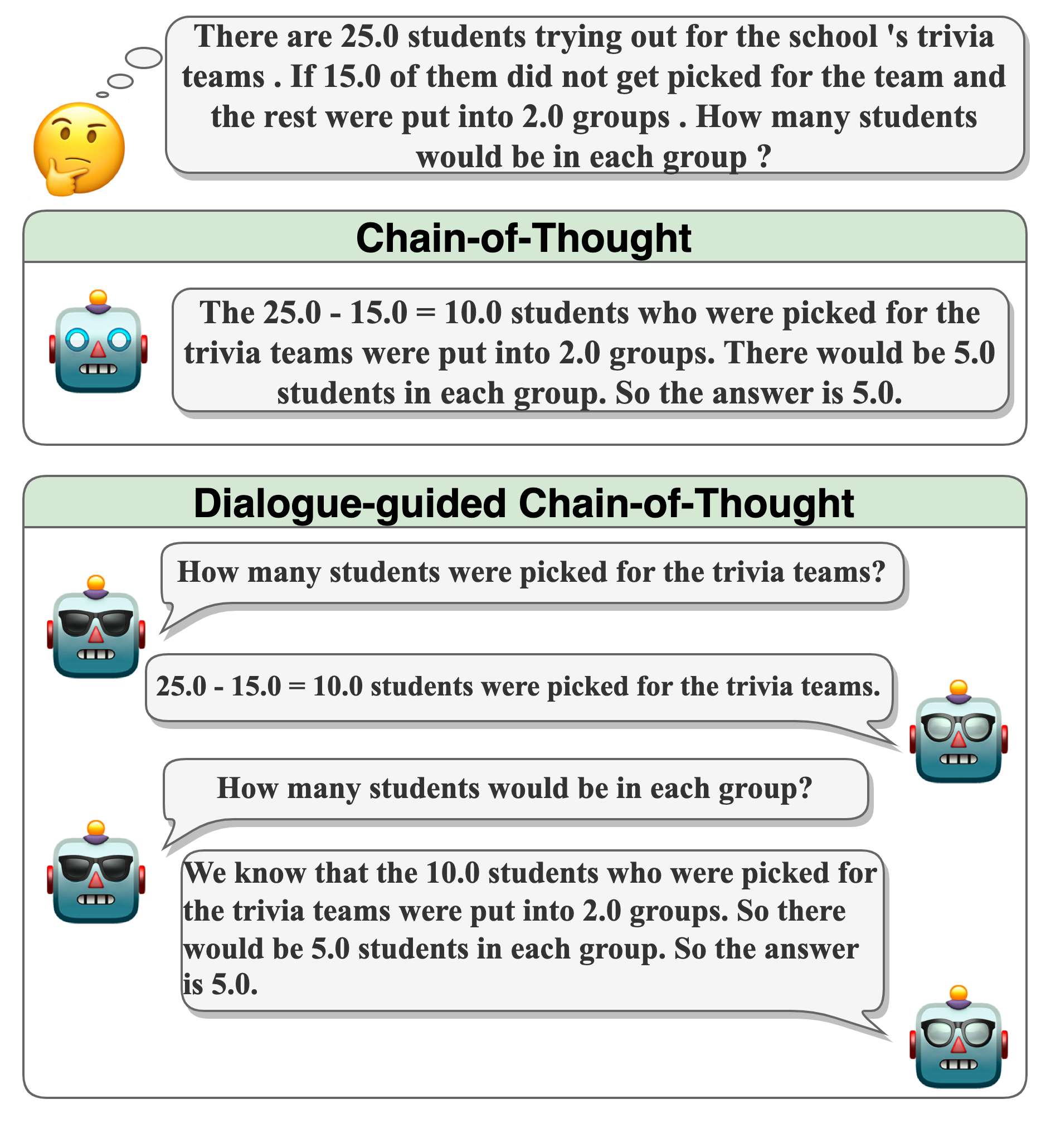}
    \caption{A comparison of DialCoT and CoT prompting~\cite{CoT_weichain}. CoT prompting guides the model to output all intermediate reasoning steps at once to obtain the final answer, while DialCoT leads the model to gradually generate intermediate reasoning steps in a dialogic format to ultimately arrive at the answer.}
    \label{fig:intro}
\end{figure}

With the advent of Chain-of-Thought~(CoT) prompting~\cite{CoT_weichain},
which encourages Large Language Models~(LLMs) to generate a series of intermediate steps 
to help get the final answer,
reasoning capabilities of LLMs
has seen significant improvement.
However,
preliminary results of \citet{CoT_weichain} have demonstrated that
CoT prompting only shows significant performance gains on LLMs~($\geq\!\!100$B),
such as LaMDA-137B~\cite{LaMDA:journals/corr/abs-2201-08239},
GPT-3 175B~\cite{GPT3:conf/nips/BrownMRSKDNSSAA20}
and PaLM-540B~\cite{PaLM:journals/corr/abs-2204-02311}.
But it is ineffective, or even detrimental,
to the performance on reasoning tasks in Smaller Language Models~($\leq\!\!10$B).
This phenomenon is explained by~\citet{CoT_weichain},
who attribute it to abilities 
such as semantic comprehension and symbolic mapping, which only manifest at larger scales.
The massive computational requirements and inference costs of LLMs 
make them unfeasible for widespread deployment.
There is a pressing community interest in figuring out 
how to further enhance the reasoning capabilities within Smaller Language Models~(SLMs).

Recent works~\cite{Teaching:journals/corr/abs-2212-08410,Teachers:journals/corr/abs-2212-10071, Specialized:journals/corr/abs-2301-12726} 
have attempted to enhance the performance of SLMs on reasoning tasks 
by fine-tuning them with training data 
generated from LLMs
that contain intermediate reasoning steps. 
However, the results have been less than optimal.
To further boost the reasoning capabilities of SLMs,
we propose \textbf{Dial}ogue-guided \textbf{C}hain-\textbf{o}f-\textbf{T}hought~(\textbf{DialCoT}),
which aims to progressively generate intermediate reasoning steps
in the dialogue format, 
instead of generating all intermediate reasoning steps at once~(as shown in Figure~\ref{fig:intro}).
Specifically, we assign to the model two roles:
\emph{Decomposer} and \emph{Solver}. 
The \emph{Decomposer} is tasked with 
breaking down the original question into a series of sub-questions.
The \emph{Solver} sequentially addresses each sub-question presented by the Decomposer, 
thereby obtaining the answer to the original question.
They utilize different instructions 
while sharing the same model parameters.
We propose three different forms of DialCoT:
1) \textbf{DialCoT-A}~(\textbf{A}ll at once), 
in which the Decomposer generates all sub-questions at once
and the Solver simultaneously provides all answers. 
2) \textbf{DialCoT-M}~(\textbf{M}ixed),
where the Decomposer generates all sub-questions at once
but the Solver sequentially delivers the answers
of the sub-questions generated by the Decomposer.
3) \textbf{DialCoT-S}~(\textbf{S}tep by step),
where both the Decomposer and Solver 
operate sequentially to generate sub-questions and their corresponding answers.
We provide a detailed comparison of the performance of the three different forms of DialCoT
in Section~\ref{sec:results}.
Furthermore,
building upon DialCoT-S,
we design \textbf{DialCoT-S-PPO},
which leverages the \textbf{P}roximal \textbf{P}olicy \textbf{O}ptimization algorithm 
to select the optimal reasoning path,
thereby further enhancing its performance in reasoning tasks.
Compared to previous methods~\cite{Teachers:journals/corr/abs-2212-10071,Specialized:journals/corr/abs-2301-12726},
our approach has two main advantages:
\begin{enumerate}
    \item 
    We transform the process of solving a complex reasoning question into 
    decomposing the question and solving a series of simpler sub-questions, 
    which reduces the task difficulty and is more suitable for SLMs.
    \item 
    By breaking down intermediate reasoning steps 
    into dialogue-formatted sub-questions and answers,
    we can use reinforcement learning more effectively 
    to choose the optimal reasoning path from various options.
\end{enumerate}
To validate the effectiveness of our approach, 
we fine-tune Flan-T5~\cite{FlanT5:journals/corr/abs-2210-11416}
using 7000 training examples from the GSM8K~\cite{GSM8K:journals/corr/abs-2110-14168} dataset
that include intermediate questions and answers.
The results surpass the latest method, SpecialFT~\cite{Specialized:journals/corr/abs-2301-12726},
by $6.2$\%.
Notably, the amount of training data we use is only $1/20$ of that used by SpecialFT.
In addition, to verify the model's generalization capability on out-of-distribution tasks, 
we also test on the MultiArith~\cite{MultiArith:conf/emnlp/RoyR15}, 
ASDiv~\cite{ASDiv:conf/acl/MiaoLS20}
and SVAMP~\cite{SVAMP:conf/naacl/PatelBG21} datasets. 
Our method achieves state-of-the-art performance compared to other baselines.


\section{Related Work}

\subsection{Chain-of-Thought Prompting}

Chain-of-Thought~(CoT),
which significantly enhances the reasoning capacities of large language models,
was originally pioneered by \citet{CoT_weichain}.
The approach focuses on augmenting few-shot examples with detailed reasoning steps, 
thereby markedly improving performance on reasoning tasks. 
Subsequent works, inspired by \citet{CoT_weichain},
have further refined the CoT methodology, 
such as Self-Consistency~\cite{SC:journals/corr/abs-2203-11171}, 
Least-to-Most prompting~\cite{LtoM:journals/corr/abs-2205-10625}, 
Dynamic Least-to-Most prompting~\cite{DLtoM:journals/corr/abs-2209-15003}, 
Self-Training~\cite{Self-training:journals/corr/abs-2210-11610}, 
Verifier~\cite{Verifier:journals/corr/abs-2206-02336}
and Tree of Thought~\cite{ToT:journals/corr/abs-2305-10601}.
The aforementioned methods primarily focus on 
improving the specific format of CoT prompting 
to better stimulate the reasoning capabilities of LLMs~($\geq\!\!100$B).
However, they are not tailored 
to augment the reasoning capabilities of SLMs~($\leq\!\!10$B).
We propose a novel method
specifically designed to enhance the performance of SLMs on reasoning tasks.

\subsection{Reasoning Enhancement in SLMs}

\citet{FlanT5:journals/corr/abs-2210-11416} observes that 
training SLMs with data 
that include intermediate reasoning steps 
can improve the reasoning capabilities of SLMs. 
Both \citet{Teaching:journals/corr/abs-2212-08410}
and \citet{Teachers:journals/corr/abs-2212-10071} 
enhance the reasoning capabilities of SLMs 
by fine-tuning them with training data, 
which includes intermediate reasoning steps generated by LLMs. 
STaR~\cite{STaR:conf/nips/ZelikmanWMG22} enables the model 
to self-improve through its own generated rationales. 
SpecialFT~\cite{Specialized:journals/corr/abs-2301-12726} employs LLMs as teacher models 
and utilizes distribution matching 
in knowledge distillation 
to transfer the reasoning capabilities from LLMs to SLMs.
Orca~\cite{mukherjee2023orca} learns to imitate the reasoning process of LLMs
from rich signals generated by LLMs,
including explanation traces,
step-by-step thought processes
and other complex instructions.
Differently,
DialCoT transforms solving complex reasoning questions 
into decomposing questions
and addressing a series of simpler questions,
significantly reducing the task difficulty.
Furthermore,
we incorporate the PPO algorithm 
to enable the model 
to choose the optimal reasoning path 
among multiple options,
thereby further enhancing
the performance in reasoning tasks.
Notably, our method does not require 
generating a large amount of training data 
with intermediate reasoning steps through LLMs. For example,
by simply fine-tuning with only 7,000 examples from the GSM8K dataset, we could achieve a remarkable enhancement in SLM performance on reasoning tasks.
\subsection{Question Decomposition}

Question decomposition is crucial for understanding and solving complex questions. 
Earlier research~\cite{Decom2012:journals/ibmrd/KalyanpurPBLC12}
uses decomposition rules based on lexico-syntactic features to facilitate question decomposition.
HSP~\cite{HSP:conf/acl/ZhangCXW19} proposes a hierarchical semantic parsing method 
based on a sequence-to-sequence model, 
which combines a question decomposer and an information extractor.
\citet{HQD:conf/emnlp/PatelMPB22} designs a human-in-the-loop question decomposition method
to improve model performance.
Least-to-Most prompting~\cite{LtoM:journals/corr/abs-2205-10625}
improves the format of CoT, 
enhancing the reasoning capabilities of LLMs 
by decomposing problems.
Self-Ask~\cite{selfask_2022measuring} explicitly asks itself
follow-up questions before answering the initial question
to perform compositional reasoning tasks.
DecomT5~\cite{DecomT5:conf/emnlp/Zhou0YR22} develops robust decomposition-based models
using distant supervision from comparable texts.
Decomposition Distillation~\cite{DecomDistill} 
learns a semantic decomposition of the original question 
into a sequence of sub-questions 
and uses it to train two models designated for question decomposition and resolution.
Compared to the aforementioned methods, 
we not only decompose the question 
but also enable the model to choose the optimal reasoning path 
through reinforcement learning methods,
thereby further enhancing the model's capability to solve complex questions.

\begin{figure*}[htbp]
    \centering
    \includegraphics[width=0.9\textwidth]{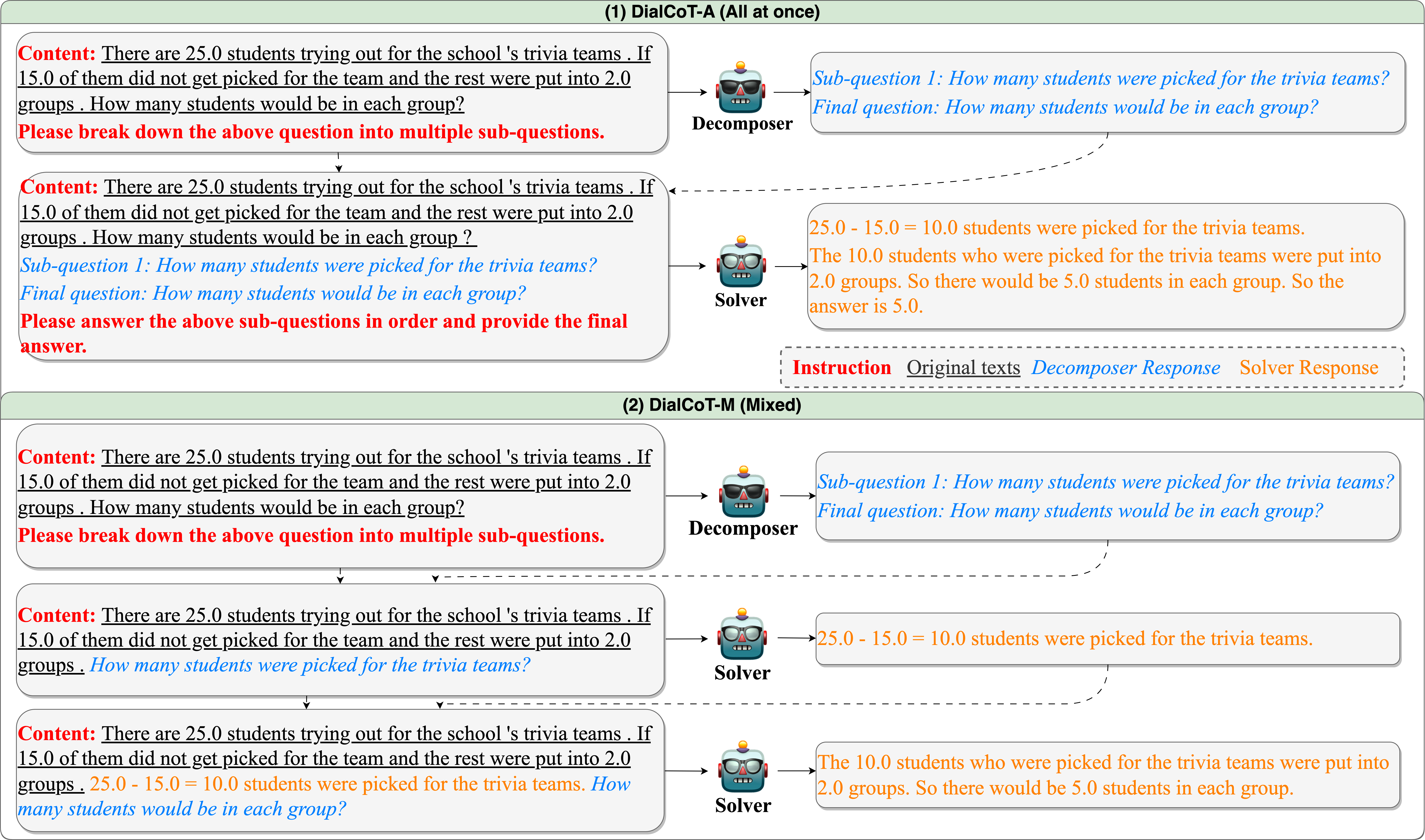}
    \caption{(1) \textbf{DialCoT-A},
    where the Decomposer generates all sub-questions at once,
    and the Solver responds to all the sub-questions in a single step.
    (2) \textbf{DialCoT-M}, 
    where the Decomposer is the same as in DialCoT-A,
    while the Solver addresses a sub-question at a single step,
    with the response being incorporated into
    the original texts to aid in solving subsequent sub-questions.
    }
    \label{fig:model}
\end{figure*}

\section{Dialogue-guided Chain-of-Thought}

We propose Dialogue-guided Chain-of-Thought (DialCoT),
which aims to decompose complex questions 
into sub-questions in a dialogue format 
and gradually guide the model 
to obtain the final answer.
Specifically, 
we introduce two roles for the model, 
namely the \emph{Decomposer} and the \emph{Solver}, 
who engage in a dialogue-based interaction.
The Decomposer is responsible for 
breaking down the original question 
into a series of simpler sub-questions, 
while the Solver sequentially answers these sub-questions.
We design distinct instructions for the Decomposer and Solver, 
followed by performing instruction tuning~\cite{wei2021InstructionTuning} on SLMs.
We first introduce three different forms of DialCoT.
Subsequently, 
we describe 
how we incorporate the Proximal Policy Optimization~(PPO) algorithm 
into DialCoT to enable the model 
to select the optimal reasoning path 
and further enhance its reasoning capabilities.

\subsection{Three Forms of DialCoT}

\begin{figure*}[!t]
    \centering
    \includegraphics[width=0.9\textwidth]{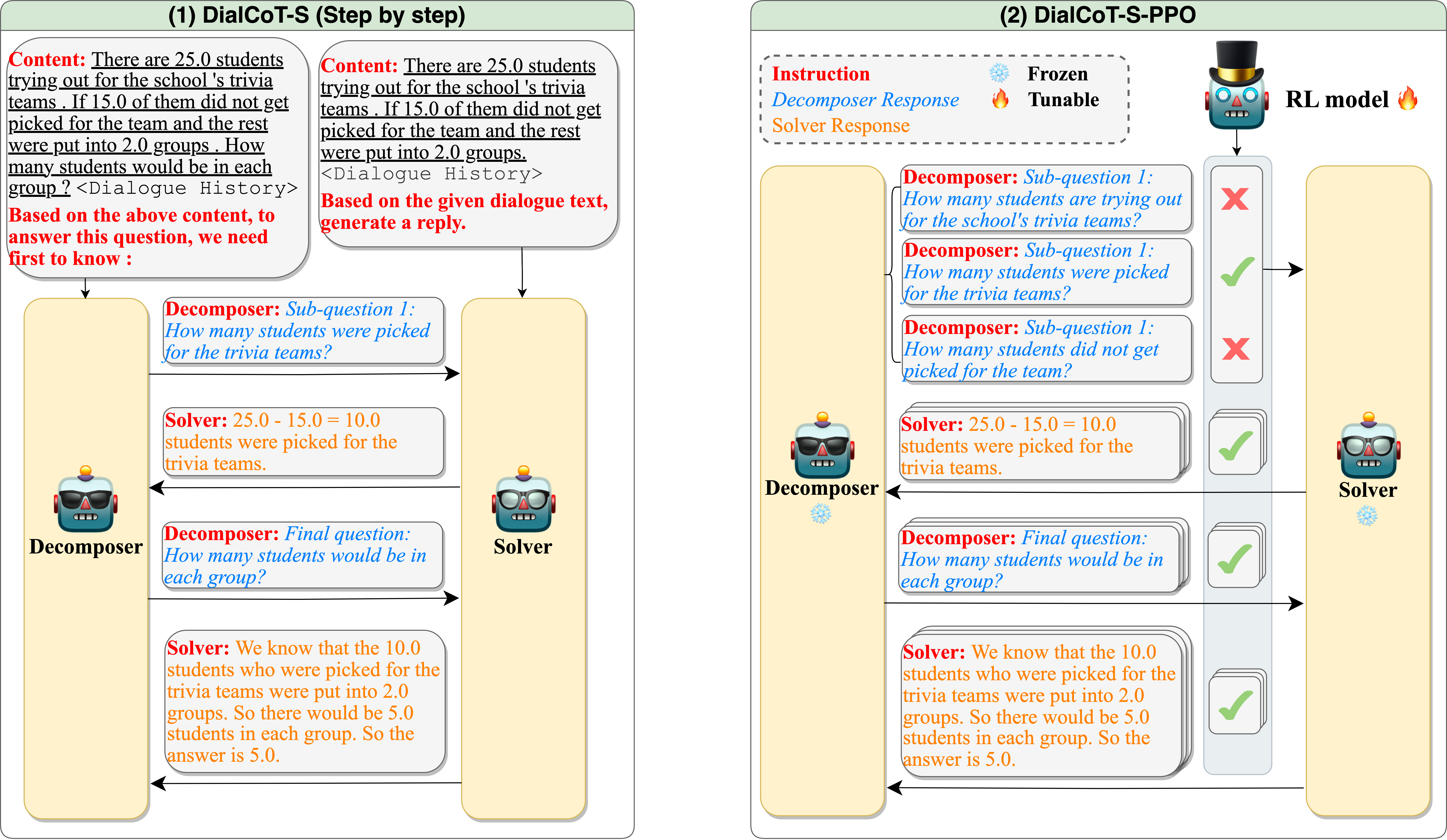}
    \caption{(1) \textbf{DialCoT-S},
    where the Decomposer presents a sub-question at a step
    and the Solver answers it. 
    Their past dialogue information is inserted into \texttt{<Dialogue History>}
    to assist in generating the answer of the final question.
    (2) An example of \textbf{DialCoT-S-PPO} solving a math word problem. 
    The policy network is used to select a response from each step of the SLM, 
    ultimately forming an optimal reasoning path and arriving at the final answer.}
    \label{fig:rl}
\end{figure*}

We propose three different dialogue forms of DialCoT, 
namely DialCoT-A, DialCoT-M, and DialCoT-S.
Specifically,
DialCoT-A aims to guide SLMs in reasoning 
through minimal dialogue turns.
DialCoT-M refines the Solver based on DialCoT-A,
further reducing the task complexity.
DialCoT-S maximally decomposes intermediate reasoning steps,
allowing it to
reference previous sub-questions and their answers 
when proposing new sub-questions.
Figure~\ref{fig:model} and Figure~\ref{fig:rl}(1) presents the overall frameworks of them.\footnote{
The specific prompt structures can be found in Table~\ref{tab:selfask} of Appendix~\ref{sec:selfask}.
}
Next, we will introduce each of these forms individually.

\paragraph{DialCoT-A~(All at once).}
We first establish an instruction for the Decomposer,
enabling it to generate all sub-questions 
in a single step.
Subsequently, 
we incorporate the generated sub-questions to the original texts
and design a new instruction for the Solver,
which allows the Solver 
to answer all sub-questions simultaneously.
Figure~\ref{fig:model}(1) displays the instructions we designed 
and examples of input/output 
when the model operates as a Decomposer and Solver.
DialCoT-A shares a similar motivation with Orca~\cite{mukherjee2023orca}, 
both striving to improve model reasoning performance 
by providing explicit reasoning paths. 
Orca represents the reasoning path 
through problem-solving steps, 
whereas our method 
exhibits the reasoning path 
via a sequence of sub-questions.

\paragraph{DialCoT-M~(Mixed).}
Upon deriving a series of sub-questions 
via the same Decomposer used in DialCoT-A,
we sequentially replace the final question 
in the original texts with these sub-questions,
which allows the Solver to address each sub-question individually. 
The Solver's response from each sub-question
is appended to the original text, 
providing contextual support for solving subsequent questions. 
Figure~\ref{fig:model}(2) presents an example of DialCoT-M 
solving a math word problem.
Compared to DialCoT-A,
DialCoT-M mitigates the task complexity 
for the Solver 
by addressing a single and simpler question 
in each step.

\paragraph{DialCoT-S~(Step by Step).}
We design new instructions 
to direct the Decomposer
to generate only a single sub-question at a step 
and the Solver to address the sub-question. 
Responses are prefixed with role identifiers 
such as ``\emph{Decomposer: }'' and ``\emph{Solver: }''.
The history of the dialogue is appended 
after the original texts,
aiding the model in answering subsequent questions
and deriving the final answer.
Figure~\ref{fig:rl}(1) displays the overall framework of DialCoT-S.
Compared to the previous two forms of DialCoT, 
DialCoT-S can reference previous sub-questions 
and their answers when generating new sub-questions.
Moreover, 
DialCoT-S is more similar to 
the traditional multi-turn dialogue format.
Therefore,
it can more effectively stimulate 
the model's multi-turn dialogue capability
to improve the model performance on reasoning tasks.

\subsection{DialCoT-S-PPO}

DialCoT-S-PPO aims to enable the model to select the optimal reasoning path
by combining DialCoT-S with the PPO algorithm, 
further improving the model's performance on reasoning tasks.
Figure~\ref{fig:rl}(2) presents an example of DialCoT-S-PPO solving a math word problem.
DialCoT-S-PPO chooses the optimal intermediate questions or answers
from the model's multiple outputs,
thus forming an reasoning path through a series of choices.
Specifically,
we first need to collect some data composed of 
states $\mathcal{S}$, 
actions $\mathcal{A}$
and rewards $\mathcal{R}$ for training the policy network $\pi_\theta$.
$\mathcal{S}$ represents the space of states of the environment,
which are the input of the policy network. Let $\mathbf{s}_t\in\mathcal{S}$ 
be a state at time $t$ defined as
\begin{align}
\label{eq:state}
    \mathbf{s}_t = [\mathbf{h}_1; \mathbf{h}_2;...;\mathbf{h}_k], \\
    [\mathbf{h}_1, \mathbf{h}_2,...,\mathbf{h}_k] = \mathtt{LM}_{\phi}(X),
\end{align}
where $X$ denotes the input text constructed via DialCoT-S,
$\mathtt{LM}_{\phi}(\cdot)$ represents the SLM after instruction tuning
and $\mathbf{h}$ denotes the last hidden state of the model's response.
We utilize beam search 
to generate the top-$k$ responses 
with the highest probability as candidates
and $\mathbf{s}_t$ is the concatenation 
of these candidates' last hidden states.
$\mathcal{A}=[0,1,...,k]$ represents the action space. 
At each time $t$,
we input $\mathbf{s}_t$ into $\pi_{\theta}$ 
to obtain the probability $\mathbf{p}$ of actions:
\begin{align}
\label{eq:action}
    \mathbf{p} = \pi_{\theta}(\mathbf{s}_t).
\end{align}
Based on the probability $\mathbf{p}$,
$\pi_\theta$ chooses an action $\mathbf{a}_t \in \mathcal{A}$,
which represents choosing the $\mathbf{a}_t$-th candidate.
During the exploration phase,
$\pi_{\theta}$ obtains $\mathbf{a}_t$ through sampling. 
In the inference phase, 
$\pi_{\theta}$ chooses the $a_t$ with the highest probability.
When the model correctly answers a sub-question,
it receives a reward $r_{m} \in [0,1]$.
$r_{m}$ is a hyperparameter
that represents the extent to 
which the model focuses on the intermediate steps.
When the model correctly answers the final question,
it receives a reward $r_{f} = 1$. 
In all other cases, the reward is $0$.
We update the policy network $\pi_{\theta}$ 
through the following objective function: \par
{\small 
\begin{align}
\label{eq:loss}
J(\theta)\!=\!\mathbb{E}_{\pi}\!\Biggl[\min&\biggl(\frac{\pi_{\theta}\left(\mathbf{a}_{t} \mid \mathbf{s}_{t}\right)}{\pi_{\theta_{\mathrm{old}}}\left(\mathbf{a}_{t} \mid \mathbf{s}_{t}\right)}A_{t}, \notag \\
&\operatorname{clip}\Bigl(\frac{\pi_{\theta}\left(\mathbf{a}_{t} \mid \mathbf{s}_{t}\right)}{\pi_{\theta_{\mathrm{old}}}\left(\mathbf{a}_{t} \mid \mathbf{s}_{t}\right)}, 1-\epsilon, 1+\epsilon\Bigr)\!A_{t}\biggr)\Biggr],
\end{align}}
where $\pi_{\theta}(\cdot)$ represents the training policy network and
$\pi_{\theta_{\mathrm{old}}}$ denotes the policy network 
that interacted with the environment to collect data.
Further information
regarding the $\operatorname{clip}(\cdot)$ and $A_{t}$ 
can be found in \citet{schulman2017PPO}.
After update the parameters of $\pi_{\theta}$,
the new parameter of $\pi_{\theta}$ is transmitted to $\pi_{\theta_{\mathrm{old}}}$.
We then repeat data collection 
and $\pi_{\theta}$ updates 
until training completion.

\section{Experiments}

\subsection{Datasets}
We consider the following four math word problem datasets:
\textbf{GSM8K}~\cite{GSM8K:journals/corr/abs-2110-14168}, \textbf{MultiArith}~\cite{MultiArith:conf/emnlp/RoyR15},
\textbf{ASDiv}~\cite{ASDiv:conf/acl/MiaoLS20} and \textbf{SVAMP}~\cite{SVAMP:conf/naacl/PatelBG21}.
The GSM8K dataset contains 7,000 training instances with intermediate questions and answers. 
In contrast to previous work~\cite{Teaching:journals/corr/abs-2212-08410,Teachers:journals/corr/abs-2212-10071,Specialized:journals/corr/abs-2301-12726},
we only fine-tune our model using these 7,000 instances, 
eliminating the need for generating additional training data 
with intermediate reasoning steps via LLMs.
Apart from evaluating our method on the GSM8K test set, 
we evaluate the model's out-of-distribution performance on three other datasets.
All datasets comprise arithmetic reasoning problems at a primary school level, 
varying by the entities they incorporate. 
This form of out-of-distribution generalization 
is typically classified as lexical-level compositional generalization~\cite{liu2021challenges}.
Following SpecialFT~\cite{Specialized:journals/corr/abs-2301-12726},
for each dataset, 
we employ 500 instances 
as the validation set, 
using the remaining instances~(800 for GSM8K, 
400 for MultiArith, 
18K for ASDiv,
500 for SVAMP) as the test set.

\subsection{Baselines}
In our experiments,
we compare our method
with some competitive baselines
which can be grouped into two categories:
(1) \emph{generic large language models}:
\textbf{code-davinci-002}~\cite{codex2021evaluating} presumably with a size of 175B or more,
\textbf{LaMDA-137B}~\cite{LaMDA:journals/corr/abs-2201-08239},
\textbf{PaLM-60B}~\cite{PaLM:journals/corr/abs-2204-02311}
and \textbf{UL2-20B}~\cite{UL2_2022unifying},
each of which exhibits strong reasoning abilities in Chain-of-Thought prompting.
(2) \emph{concurrent works enhancing SLMs' reasoning capabilities}:
\textbf{CoT-FT}~\cite{wei2021InstructionTuning}
directly employs the 7000 CoT training instances from the GSM8K dataset 
to perform instruction tuning,
which is a vanilla approach to enhancing the reasoning capabilities of SLMs.
\textbf{DecomDistill}~\cite{DecomDistill} is a decomposition-based method,
which learns a semantic decomposition of the original problem 
into a sequence of sub-problems through LLMs.
Both \citet{Teaching:journals/corr/abs-2212-08410}
and \citet{Teachers:journals/corr/abs-2212-10071}
fine-tune SLMs by generating training data 
with intermediate reasoning steps through LLMs.
\textbf{SpecialFT}~\cite{Specialized:journals/corr/abs-2301-12726} employs LLMs as teacher models 
and utilizes distribution matching in knowledge distillation
to transfer the reasoning capabilities from LLMs to SLMs. 
It is noted that 
SpecialFT uses 130K training instances with intermediate reasoning steps generated by LLMs,
which is nearly twenty times the size of our training set.

\begin{table}[!t]
	\begin{center}
	\resizebox{1.0\columnwidth}{!}{
		\begin{tabular}{cc}
			\hline
			\textbf{Hyperparameters} & \textbf{Scope}\\
            \hline
			learning rate & \{$1e-4, \textbf{3e-4}, 5e-4, 1e-3$\} \\
			batch size & \{$1024, 2048, \textbf{4096}$\} \\
			$\epsilon$ & \{$0.1, \textbf{0.2}, 0.3$\} \\
			$k$ & \{$2, \textbf{3}, 4, 5, 6$\} \\
			$r_m$ & \{$0.1, 0.2, \textbf{0.3}, 0.4, 0.5$\} \\
			\hline
		\end{tabular}
		}
		\caption{The searching scope for the hyperparameters of 
                    the proximal policy optimization algorithm. 
                    We highlight the best settings in bold.}
		\label{tab:hyper}
	\end{center}
\end{table}

\subsection{Implementation}

We consider using FlanT5-XL~(3B)/XXL~(11B) 
as the backbone of our model.
The Decomposer and Solver utilize different instructions~(as shown in Figure~\ref{fig:model})
but share the same model parameters.
Following \citet{FlanT5:journals/corr/abs-2210-11416},
we fine-tune the model for 50 epochs 
with the batch size $4096$ and the learning rate $5e-4$.
For the PPO algorithm,
we use three feed-forward layers as the policy network
and set the number of hidden units to 1024.
Moreover,
we use the grid search to 
find the best hyperparameters.
The details of grid search are shown in Table~\ref{tab:hyper}.
As a result, 
we set the learning rate as $3e-4$ and batch size as $4096$ for the policy network.
We also set $\epsilon$ to $0.2$, $k$ to $3$
and $r_m$ to $0.3$.
During the stage of optimizing the policy network,
we freeze the backbone parameters.
All baseline results except CoT-FT~\cite{wei2021InstructionTuning}
are recorded in SpecialFT~\cite{Specialized:journals/corr/abs-2301-12726}.
For CoT-FT and our method,
we keep the experimental setup consistent with other baselines.
We run all the experiments on eight NVIDIA Tesla A100 GPU.

\begin{table*}[t]
	\begin{center}
	\resizebox{2.08\columnwidth}{!}{
		\begin{tabular}{lrrccccc}
			\hline
			\hline
                \textbf{Methods}&\textbf{Backbone}&\#\textbf{Params.}&\textbf{GSM8K}&\textbf{MultiArith}&\textbf{ASDiv}&\textbf{SVAMP}&\textbf{Average} \\
                \hline
                \multicolumn{8}{l}{\textit{\textbf{Generic Large Language Models}}}\\
			code-davinci-002~\cite{codex2021evaluating}
                & 
                & $\geq 175$B
                & $63.1$
                & $95.8$
                & $80.4$
                & $76.4$
			& $78.9$ \\
			LaMDA~\cite{LaMDA:journals/corr/abs-2201-08239}
                & 
                & $137$B
                & $14.8$
                & $45.0$
                & $46.6$
                & $37.5$
			& $36.0$ \\
			PaLM~\cite{PaLM:journals/corr/abs-2204-02311}
                & 
                & $60$B
                & $29.9$
                & $75.0$
                & $61.9$
                & $46.7$
			& $53.2$ \\
			UL2~\cite{UL2_2022unifying}
                & 
                & $20$B
                & $4.4$
                & $-$
                & $16.9$
                & $12.5$
			& $-$ \\
                \hline
                \multicolumn{7}{l}{\textit{\textbf{Concurrent Works to Boosting SLM Reasoning}}}\\
                DecomDistill$^\dag$\cite{DecomDistill}
                & GPT
                & $7$B
                & $21.0$
                & $-$
                & $-$
                & $-$
			& $-$ \\
			\citet{Teaching:journals/corr/abs-2212-08410}$^\dag$
                & T5-XXL
                & $11$B
                & $21.9$
                & $-$
                & $42.1$
                & $-$
			& $-$ \\
			\citet{Teachers:journals/corr/abs-2212-10071}$^\dag$
                & GPT
                & $6$B
                & $6.8$
                & $33.3$
                & $-$
                & $-$
			& $-$ \\
                \hdashline
			CoT-FT~\cite{wei2021InstructionTuning}
                & FlanT5-XL
			& $3$B
                & $13.5$
                & $24.0$
                & $20.7$
                & $17.7$
			& $19.0$ \\
			SpecialFT$^\dag$~\cite{Specialized:journals/corr/abs-2301-12726}
                & FlanT5-XL
			& $3$B
                & $22.4$ 
                & $42.3$
                & $28.4$
                & $23.8$
			& $29.3$ \\
                DialCoT-A~(All at once)
                & FlanT5-XL
			& $3$B
                & $20.3$			
                & $40.3$
                & $24.6$
                & $21.3$
			& $26.6$ \\
                DialCoT-M~(Mixed)
                & FlanT5-XL
			& $3$B
                & $22.9$			
                & $43.1$
                & $27.1$
                & $23.2$
			& $29.1$ \\
                DialCoT-S~(Step by Step)
                & FlanT5-XL
			& $3$B
                & $24.3$			
                & $45.7$
                & $29.3$
                & $25.5$
			& $31.2$ \\
                DialCoT-S-PPO
                & FlanT5-XL
			& $3$B
                & $25.6$			
                & $46.9$
                & $30.7$
                & $27.1$
			& $32.6$ \\
                \hdashline
			CoT-FT~\cite{wei2021InstructionTuning}
                & FlanT5-XXL
			& $11$B
                & $16.1$
                & $51.7$
                & $36.5$
                & $39.7$
			& $36.0$ \\
			SpecialFT$^\dag$~\cite{Specialized:journals/corr/abs-2301-12726}
                & FlanT5-XXL
			& $11$B
                & $27.1$ 
                & $63.0$ 
                & $37.6$ 
                & $35.6$
			& $40.8$ \\
                DialCoT-A~(All at once)
                & FlanT5-XXL
			& $11$B
                & $21.7$			
                & $57.1$
                & $32.5$
                & $34.2$
			& $36.4$ \\
                DialCoT-M~(Mixed)
                & FlanT5-XXL
			& $11$B
                & $30.5$			
                & $63.9$
                & $38.2$
                & $37.7$
			& $42.6$ \\
                DialCoT-S~(Step by Step)
                & FlanT5-XXL
			& $11$B
                & $35.2$			
                & $65.7$
                & $39.3$
                & $40.3$
			& $45.1$ \\
                \textbf{DialCoT-S-PPO}
                & FlanT5-XXL
			& $11$B
                & \textbf{$\mathbf{37.1}$}			
                & \textbf{$\mathbf{68.1}$}
                & \textbf{$\mathbf{40.9}$}
                & \textbf{$\mathbf{41.7}$}
			& \textbf{$\mathbf{47.0}$} \\
			\hline
			\hline
		\end{tabular}
		}
	\end{center}
	\caption{Accuracy~(\%) of various methods on four reasoning tasks.
                $^\dag$ indicates that the method employs additional training data 
                with intermediate reasoning steps generated via LLMs,
                where SpecialFT uses nearly 20 times the training data of our method.
                We highlight the best results on SLMs~($\sim\!10$B) in bold.}
	\label{tab main experiment}
	
\end{table*}

\subsection{Results}
\label{sec:results}

Table~\ref{tab main experiment} shows the performance of various methods 
on four arithmetic reasoning datasets.
First, we discuss the results for DialCoT-A, DialCoT-M, and DialCoT-S.
Then, we compare our method with other baselines
to demonstrate its superiority.
Finally, 
we validate the effectiveness of incorporating the PPO algorithm 
based on DialCoT-S through an ablation study.

\paragraph{Discussion of Three DialCoT Variants.}
Compared to CoT-FT,
all three forms of DialCoT outperform on the four reasoning tasks,
demonstrating their effectiveness 
in enhancing the reasoning capacities of SLMs.
More specifically, 
DialCoT-M performs better than DialCoT-A.
This indicates that 
SLMs lack the capability 
to decompose a reasoning problem and answer it all at once.
DialCoT-M addresses only one sub-question
at a single step, 
which reduces the task difficulty 
and makes it more suitable for SLMs.
DialCoT-S, 
in comparison to DialCoT-M, 
shows greater performance gains, 
which can be attributed to two factors: 
(1) DialCoT-S obtains more intermediate information 
before generating sub-questions. 
(2) DialCoT-S more effectively 
stimulate the model's multi-turn dialogue capabilities 
to boost its reasoning performance.\footnote{
A more detailed discussion of the three DialCoT variants can be found in Appendix~\ref{detailed_discussion}.
}

\paragraph{Comparison between DialCoT and Baselines.}
From the Table~\ref{tab main experiment},
we observe that
DialCoT-S-PPO attains state-of-the-art results on SLMs.
Specifically,
when using FlanT5-XXL as the backbone, 
DialCoT-S-PPO improves the average performance across the four datasets 
by $6.2\%$ compared to SpecialFT.
Notably,
the training data we used is only $1/20$ of what SpecialFT used,
which clearly demonstrates that our method 
is very effective in improving reasoning capabilities of SLMs.
On the other hand, 
when compared with LLMs,
all variations of DialCoT~(i.e., DialCoT-A/M/S/PPO)
outperform LaMDA-137B on average across the four datasets, 
despite the parameters of our approach are merely $1/12$ of LaMDA's.
This further substantiates the superiority of our approach.
While there is still a noticeable gap 
when compared to code-davinci-002, 
our experimental results demonstrate that 
there is potential for SLMs 
to achieve LLM-level reasoning capabilities 
via appropriate fine-tuning methods.\footnote{The detailed experimental comparison between DialCoT and SelfAsk can be found in Appendix~\ref{sec:selfask}.}

\paragraph{Ablations.}
The ablation study is conducted 
to demonstrate the effectiveness of 
incorporating the PPO algorithm based on DialCoT-S.
Compared to DialCoT-S with FlanT5-XXL,
DialCoT-S-PPO achieves an improvement of nearly $2\%$,
confirming the effectiveness of employing the PPO algorithm 
for selecting the optimal reasoning path. 
Additionally, 
we observe that when using FlanT5-XL as the backbone,
the performance gain brought by the PPO algorithm is $1.4\%$,
which is lower than the performance on FlanT5-XXL. 
This might be attributed to 
the lower diversity in the multiple replies 
generated by the smaller model.

\subsection{Analysis}

\begin{figure}[!t]
    \centering
    \includegraphics[width=0.48\textwidth]{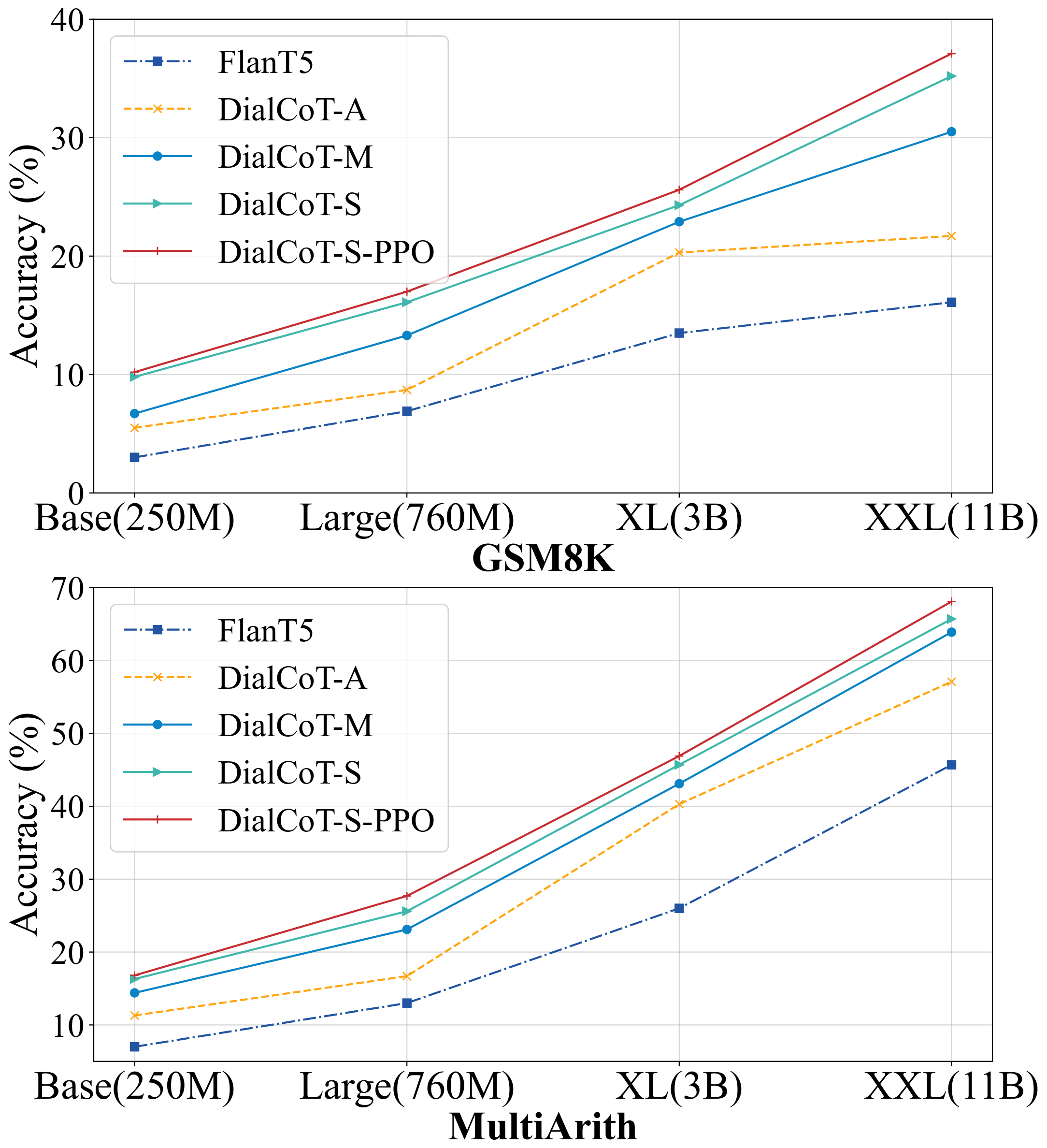}
    \caption{Results 
            under different model sizes
            on the GSM8K and MultiArith datasets. 
            FlanT5 indicates the results of using Few-shot CoT~\cite{CoT_weichain}
            on the backbone~\cite{FlanT5:journals/corr/abs-2210-11416}.
            Our methods achieve performance improvements across all model sizes.}
    \label{fig:modelsize}
\end{figure}

\paragraph{Different Model Size.}

We extend our method to smaller backbones, 
including FlanT5-Base~(250M) and FlanT5-Large~(760M),
on the GSM8K and MultiArith datasets.
Our experimental results are illustrated in Figure~\ref{fig:modelsize}.
In comparison to the original FlanT5, 
our approach improves the performance of the model on reasoning tasks 
across different model sizes, 
affirming the effectiveness of our method for varying model sizes.
Notably, 
we observe that our method yields larger performance gains 
on larger model sizes,
which is similar to
the results of \citet{FlanT5:journals/corr/abs-2210-11416}.
This could be attributed to 
the stronger capabilities 
that larger models obtain during pre-training, 
making them more readily stimulated.

\begin{figure}[!t]
    \centering
    \includegraphics[width=0.48\textwidth]{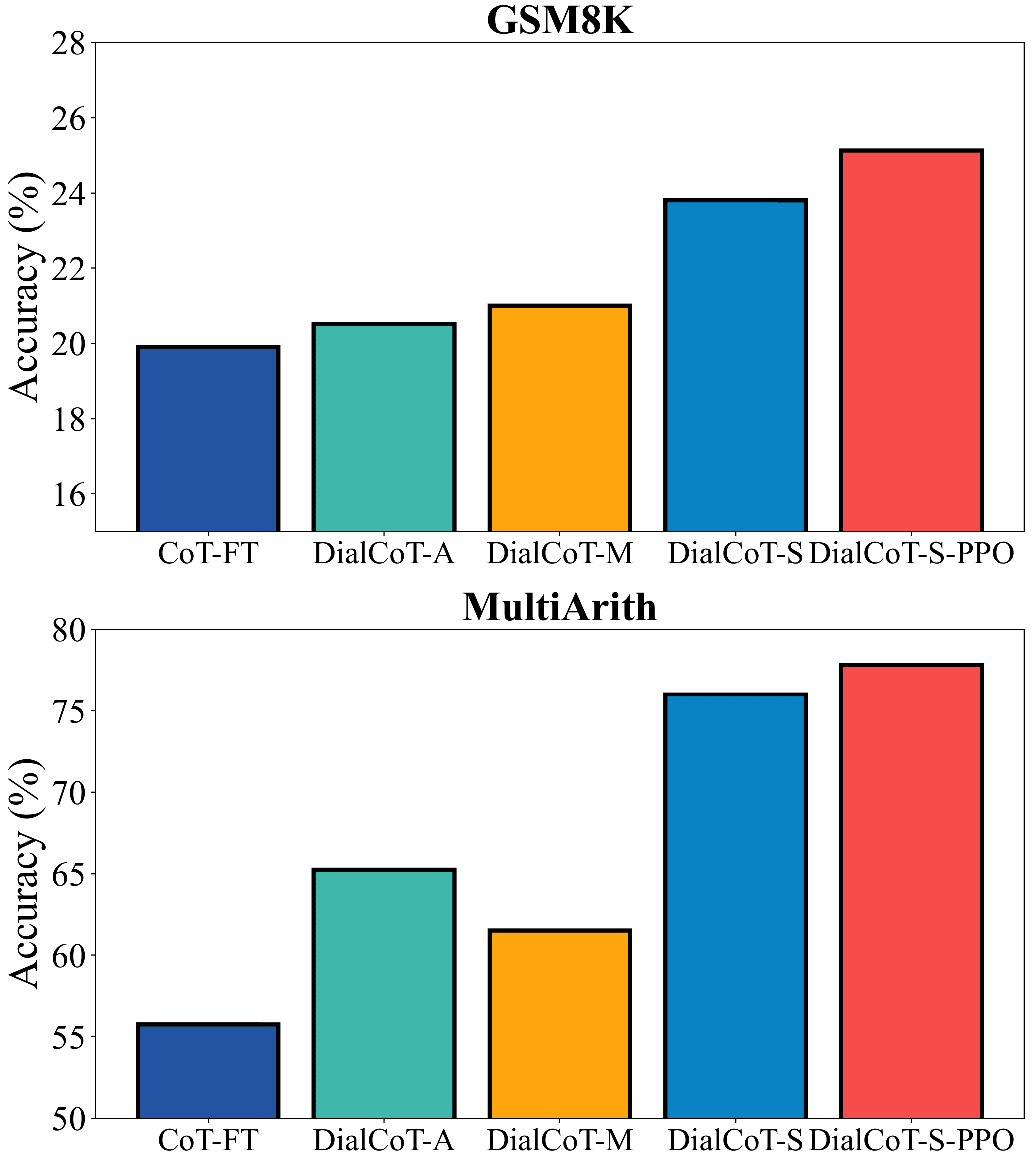}
    \caption{Results using LLaMA-7B~\cite{touvron2023llama}
    as the backbone
    on the GSM8K and MultiArith datasets.
    Our method achieves
    significant performance gains 
    on decoder-only LMs.}
    \label{fig:backbone}
\end{figure}

\paragraph{Different Model Architectures.} 
To evaluate the generalizability of DialCoT 
across LMs with varying architectures,
in addition to the encoder-decoder LM~(e.g., FlanT5),
we conduct experiments
using the decoder-only LM~(e.g., LLaMA-7B~\cite{touvron2023llama})
as the backbone of our method
on the GSM8K 
and MultiArith datasets.
The results are illustrated in Figure~\ref{fig:backbone}.
As can be seen from the figure, 
all of our methods achieve 
significant performance gains compared to CoT-FT,
especially DialCoT-S and DialCoT-S-PPO.
This demonstrates that 
our approach is applicable to 
SLMs with various architectures, 
not merely effective on encoder-decoder LMs. 
Moreover,
we observe that
DialCoT-A performs better than DialCoT-M
on the MultiArith dataset,
which is different from the results 
based on FlanT5~(as shown in Figure~\ref{fig:modelsize}).
This suggests that 
the most suitable form of DialCoT may differ for different SLMs,
which could potentially be related to 
model architecture and pre-training corpora.


\begin{figure}[!t]
    \centering
    \includegraphics[width=0.48\textwidth]{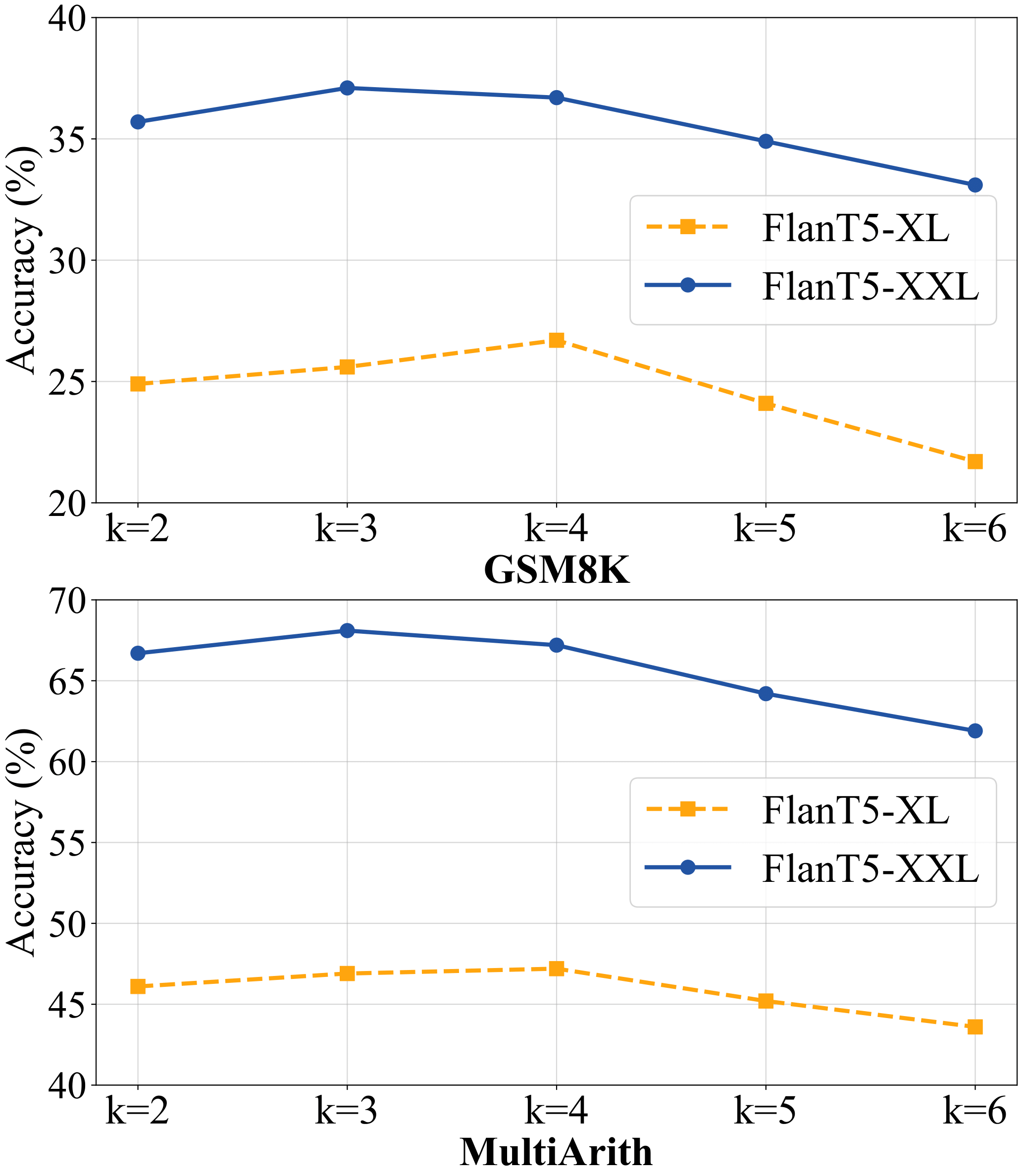}
    \caption{The effect of hyperparameter $k$ 
            on the model performance 
            on the GSM8K and MultiArith datasets. }
    \label{fig:hyper}
\end{figure}

\paragraph{Effect of Hyperparameter $k$.}
We set the model 
to output the top-$k$ responses 
with the highest probability 
in each step of dialogue
through beam search. 
In other words, 
$k$ represents the size of the action space, 
indicating that we can select the optimal response
from $k$ different responses in each step. 
Figure~\ref{fig:hyper} illustrates 
the effect of $k$ on the model performance 
on the GSM8K and MultiArith datasets.
Specifically,
the model can achieve optimal performance when $k$ is set to 3 or 4.
When $k$ is too small, 
the number of reasoning paths we can choose from is too limited,
preventing us from achieving optimal performance.
Conversely, when $k$ is too large,
the space of available reasoning paths is too large,
which may introduce noise 
and make it difficult for the model to learn to select the optimal reasoning path.

\section{Conclusion}
In this paper, 
we explored strategies to boost the reasoning capabilities of SLMs 
and proposed DialCoT, 
which aims to generate intermediate reasoning steps 
in a dialogue format leading to the final answer.
Specifically, we designed two roles for the model,
namely Decomposer and Solver. 
The Decomposer is responsible for breaking down questions 
into multiple sub-questions, 
while the Solver is tasked to address the sub-questions.
They engage in dialogue to arrive at the final answer.
We introduced three different dialogue formats: 
DialCoT-A~(All at once), 
DialCoT-M~(Mixed)
and DialCoT-S~(Step by step).
Furthermore, 
we incorporated the PPO algorithm 
into DialCoT-S to enable the model 
to choose the optimal reasoning path 
among multiple options, 
thereby further improving its performance on reasoning tasks. 
We conducted extensive experiments on four arithmetic reasoning datasets
and the experimental results demonstrate the effectiveness of our method.
Future work primarily involves 
extending our method to other types of reasoning tasks, 
such as commonsense reasoning and symbolic reasoning.
In addition, 
we will explore other decomposition methods
or other reinforcement learning methods 
to optimize the reasoning paths of SLMs.

\section*{Limitations}

We conduct experiments on four arithmetic reasoning tasks, 
demonstrating the effectiveness of DialCoT. 
However, 
as our reward pattern is specifically 
designed for arithmetic reasoning, 
modifications are necessary 
to apply our method to commonsense or symbolic reasoning. 
This presents a limitation to the broader applicability of our method. 
We plan to extend DialCoT to a wider range of reasoning tasks in the future. 
On the other hand,
DialCoT specifically focuses on 
enhancing the reasoning capabilities of SLMs. 
Due to constraints on computational resources, 
we do not conduct experiments on larger scale language models~($\geq20$B),
thus the applicability of our method for LLMs remains undetermined. 
We will further explore the performance of DialCoT 
on larger scale language models in future research.

\section*{Ethics Statement}

The proposed method has no obvious potential risks.
All the scientific artifacts used/created are properly cited/licensed, 
and the usage is consistent with their intended use. 
Also, 
we open up our codes and hyper-parameters 
to facilitate future reproduction without repeated energy cost.

\section*{Acknowledgements}
This work has been supported 
by the National Natural Science Foundation of China under Grant No.U1911203, 
the National Natural Science Foundation of China under Grant No.62377012
and
Fundamental Research Funds for the Central Universities under grant number
YBNLTS2023-015.

\bibliography{custom}
\bibliographystyle{acl_natbib}

\clearpage
\appendix

\section{Comparison between DialCoT and SelfAsk}
\label{sec:selfask}

We add further discussion 
regarding the comparison with SelfAsk. 
Self-Ask~\cite{selfask_2022measuring} explicitly asks itself
follow-up questions before answering the initial question
to perform compositional reasoning tasks.
Please refer Table~\ref{tab:selfask} 
for the comparison between SelfAsk and our methods.

From the table, we see that:
(1) SelfAsk is designed for in-context learning method 
without fine-tuning, whereas DialCoT is a finetune-based method.
(2) Even if fine-tuning can be applied to SelfAsk theoretically, 
how to format the fine-tuning is still un-explored. 
We offer a novel way to leverage two tailored tasks 
(problem decomposition and problem solving) 
through fine-tuning task-oriented instructions on the same model (SLM).
Therefore,
we believe the performance gain 
mainly come from the way of 
decomposing and solving sub-questions 
through fine-tuned model with tailored instructions. 
The additional instructions naturally come along with the proposed solution.
They will instruct the model to play dedicated role during the instruction fine-tuning. 
However, we don’t think they are the major factors.

Moreover, 
we conduct additional experiments (SelfAsk) on the GSM8K dataset using Flan-T5-XXL.
The results are illustrated in Table~\ref{tab:selfask_finetune}.
From the table, 
we can draw the following conclusions:

Firstly, 
in the fine-tuning setting, 
SelfAsk improves on Standard CoT by 5.5\% 
with all-at-once finetuning and by 13.2\% with sequential finetuning.
We believe these improvements stem from problem decomposition. 
Furthermore, SelfAsk with sequential finetuning improves by 7.7\% 
compared to SelfAsk with all-at-once finetuning. 
This once again confirms our previous conclusion
that decomposing problems sequentially 
is more effective than decomposing them all at once.
Moreover, DialCoT-S improves by 5.9\% 
over SelfAsk with sequential finetuning. 
We attribute this additional improvement 
to fine-tuning with different instructions 
tailored for specific tasks.
Compared to SelfAsk with sequential finetuning,
DialCoT-S has clearer and more independent instructions 
for both problem decomposition and problem solving.

Secondly, 
compared to finetune-based methods, 
methods without fine-tuning perform poorly, 
indicating that fine-tuning is crucial for SLMs. 
At the same time,
we found that both SelfAsk and DialCoT-S experience a performance drop 
in the setting compared to Standard CoT. 
This could be either because Flan-T5 was trained 
with some standard-CoT-formatted training data or
due to the weaker instruction-following capabilities of SLMs, 
where complex instructions increase the task difficulty.

For encoder-decoder structure, 
i.e., T5, 
the difference between SelfAsk-A and SelfAsk-S is significant, 
due to the fact of bidirectional attention within the input.
For decoder-only structure,
the difference is indeed very subtle. 
Table~\ref{tab:selfask_llama} shows the results of additional experiments (SelfAsk) 
on the GSM8K using LLaMA-7B.
The SelfAsk-A outputs all intermediate questions,
answers and connecting words between them such as ``Follow up''
and ``Intermediate answer'',
while the sequential finetuning is focused on 
outputting intermediate question or answer, 
and does not include the loss of connecting words, 
encouraging the model to focus more 
on most important part of the learning. 
We speculate this subtlety brings the improvement.

\section{Fine-grained Analysis on Each Sub-step}
\label{app:sub}

We select samples that 
require decomposition into 
three sub-questions 
from all test sets 
and report the accuracy 
for each sub-question in Table~\ref{tab:substep}.
As shown in the table,
as the number of steps increases,
i.e., 
as the complexity of the questions rises, 
our method shows
a notable performance improvement ($7\%$) 
compared to CoT-FT. 
On the other hand,
although our method's performance 
does decline as question complexity increases,
the rate of decline 
is significantly slower 
compared to vanilla CoT. 
When comparing DialCoT-S with DialCoT-S-PPO,
It is evident that 
step-level PPO significantly improves the model's performance 
on reasoning tasks at every step.

\begin{table*}[htbp]
\centering
\resizebox{2\columnwidth}{!}{
\begin{tabular}{l p{10cm} l l}
\toprule
Methods & Prompt Structure & All-at-once? & ICL or FT \\
\midrule
SelfAsk & \textit{Input:} Original question \textbar\ \textit{Output:} Sub-question 1 + Intermediate step 1 + ... + Final answer & All-at-once & In-context Learning \\
\midrule
\multirow{2}{*}{DialCoT-A} & \textbf{Decomposer:} \textit{Input:} Original question \textbar\ \textit{Output:} Sub-questions & \multirow{2}{*}{All-at-once} & \multirow{2}{*}{Fine-tuning} \\
& \textbf{Solver:} \textit{Input:} Original question + Sub-questions \textbar\ \textit{Output:} Intermediate steps + Final answer & & \\
\midrule
\multirow{5}{*}{DialCoT-M} & \textbf{Decomposer:} \textit{Input:} Original question \textbar\ \textit{Output:} Sub-questions & \multirow{5}{*}{\makecell[l]{Output intermediate \\ steps sequentially}} & \multirow{5}{*}{Fine-tuning} \\
& \textbf{Solver:} \textit{Input:} Original question + Sub-question 1 \textbar\ \textit{Output:} Intermediate step 1 & & \\
& \textbf{Solver:} \textit{Input:} Original question + Intermediate step 1 + Sub-question 2 \textbar\ \textit{Output:} Intermediate step 2 & & \\
& ... & & \\
& \textbf{Solver:} \textit{Input:} Original question + Intermediate steps + Final question \textbar\ \textit{Output:} Final answer & & \\
\midrule
\multirow{8}{*}{DialCoT-S} & \textbf{Decomposer:} \textit{Input:} Original question \textbar\ \textit{Output:} Sub-question 1 & \multirow{8}{*}{\makecell[l]{Output sub-questions \\ and intermediate steps \\ alternately}} & \multirow{8}{*}{Fine-tuning} \\
& \textbf{Solver:} \textit{Input:} Original question + Sub-question 1 \textbar\ \textit{Output:} Intermediate step 1 & & \\
& \textbf{Decomposer:} \textit{Input:} Original question + Sub-question 1 + Intermediate step 1 \textbar\ \textit{Output:} Sub-question 2 & & \\
& \textbf{Solver:} \textit{Input:} Original question + Sub-question 1 + Intermediate step 1 + Sub-question 2 \textbar\ \textit{Output:} Intermediate step 2 & & \\
& ... & & \\
& \textbf{Decomposer:} \textit{Input:} Original question + Sub-questions + Intermediate steps \textbar\ \textit{Output:} Final question & & \\
& \textbf{Solver:} \textit{Input:} Original question + Sub-questions + Intermediate steps + Final question \textbar\ \textit{Output:} Final answer & & \\
\bottomrule
\end{tabular}}
\caption{Detailed Comparison between DialCoT and SelfAsk. The table omits the specific details of the prompts and instead focuses on the input-output formats of the prompts. DialCoT is fine-tuned using distinct instructions tailored for two specialized tasks (Decomposer and Solver) on a shared model.}
\label{tab:selfask}
\end{table*}

\begin{table}[ht]
\centering
\begin{tabular}{lll}
\hline
Method         & Finetune or not           & GSM8K \\
\hline
Standard CoT   & finetune                  & 16.1  \\
SelfAsk        & all at once, finetune     & 21.6  \\
SelfAsk        & sequencially, finetune    & 29.3  \\
DialCoT-S      & finetune                  & 35.2  \\
Standard CoT   & without finetune          & 12.7  \\
SelfAsk        & without finetune          & 11.3  \\
DialCoT-S      & without finetune          & 10.9  \\
\hline
\end{tabular}
\caption{Accuracy (\%) of various methods on the GSM8K dataset. The fine-tuning dataset used for all experiments is the GSM8K training set. The prompt structures corresponding to the two different fine-tuning methods for SelfAsk are shown in Table~\ref{tab:selfask_type}.}
\label{tab:selfask_finetune}
\end{table}

\begin{table*}[ht]
\centering
\resizebox{2.08\columnwidth}{!}{
\begin{tabular}{llp{15cm}} 
\hline
Method & Fine-tuning Method & Prompt Structure \\
\hline
SelfAsk & all at once & \textit{Input:} Original question \textbar{} \textit{Output:} \textbf{Follow up:} Sub-question 1 \newline \textbf{Intermediate answer:} Intermediate step 1 \newline \textbf{Follow up:} Sub-question 2 \newline \textbf{Intermediate answer:} Intermediate step 2 + ... + \textbf{Follow up:} Final question \newline \textbf{Intermediate answer:} Final answer \\
\hline
SelfAsk & sequencially & \textit{Input:} Original question \newline \textbf{Follow up:} \textbar{} \textit{Output:} Sub-question 1 \\
& & \textit{Input:} Original question \newline Follow up: Sub-question 1 \newline \textbf{Intermediate answer:} \textbar{} \textit{Output:} Intermediate step 1 \\
& & \textit{Input:} Original question \newline Follow up: Sub-question 1 \newline Intermediate answer: Intermediate step 1 \newline \textbf{Follow up:} \textbar{} \textit{Output:} Sub-question 2 \\
& & ... \\
& & \textit{Input:} Original question \newline Follow up: Sub-question 1 \newline Intermediate answer: Intermediate step 1 + ... + \textbf{Follow up:} \textbar{} \textit{Output:} Final question \\
& & \textit{Input:} Original question \newline Follow up: Sub-question 1 \newline Intermediate answer: Intermediate step 1 + ... + Follow up: Final question \newline \textbf{Intermediate answer:} \textbar{} \textit{Output:} Final answer \\
\hline
\end{tabular}}
\caption{Prompt structures corresponding to the two different fine-tuning methods for SelfAsk.}
\label{tab:selfask_type}
\end{table*}

\begin{table}[ht]
\centering
\begin{tabular}{lr}
\hline
Method & GSM8K \\
\hline
SelfAsk + All-at-once finetuning & 21.00 \\
\hline
SelfAsk + Sequential finetuning & 22.95 \\
\hline
\end{tabular}
\caption{Accuracy (\%) of SelfAsk using LLaMA-7B on the GSM8K dataset.}
\label{tab:selfask_llama}
\end{table}

\begin{table}[!t]
\centering
\resizebox{1\columnwidth}{!}{
\begin{tabular}{cccc}
\hline
Method & first step & second step & final step\\
\hline
CoT-FT & 55.2 & 41.9 & 34.7 \\
\hline
DialCoT-S & 60.7 & 48.8 & 43.8 \\
\hline
DialCoT-S-PPO & 63.9 & 51.2 & 45.3 \\
\hline
\end{tabular}}
\caption{Accuracy (\%) of various methods using FlanT5-XXL on the GSM8K dataset for each sub-step. CoT-FT refers to the results obtained by
replacing the original question 
with different step-based sub-questions. }
\label{tab:substep}
\end{table}

\section{Comparison of Inference Speed}

In terms of the time cost for reasoning, 
the inference speed of our method 
is comparable to that of SelfAsk~\cite{selfask_2022measuring},
as both need to 
generate sub-questions and answers.
In terms of the number of iterations, 
DialCoT-A requires two iterations, 
which is comparable to 
Zero-shot CoT~\cite{kojima2022zerocot},
while DialCoT-M and DialCoT-S 
require approximately three and six iterations 
respectively. 
Overall, 
in our methods,
DialCoT-A has the fastest inference speed, 
followed by DialCoT-M, 
and finally DialCoT-S. 
The trade-off between
inference time and performance needs 
to be considered. 
If faster inference speed is required, 
one can opt for DialCoT-A, 
at the expense of some performance loss. 
Conversely, 
if better performance is the priority, 
DialCoT-S can be chosen, 
although this would come at the cost of 
increased inference time.

\section{Detailed Discussion of Three DialCoT Variants}
\label{detailed_discussion}

From Table~\ref{tab:selfask},
we can clearly see the differences 
in the prompt structures of different methods. 
Combined with the experimental results from Table~\ref{tab main experiment},
we can draw the following conclusions:
(1) In a side-by-side comparison between standard CoT and DialCoT-A, 
it becomes evident that DialCoT-A employs self-generated sub-problems 
as a strategic guide for formulating intermediate steps and solutions. 
The enhanced performance of DialCoT-A over standard CoT 
implies that self-generated navigation through 
these sub-problems can significantly enhance 
the reasoning capability of Smaller Language Models (SLMs).
(2) When comparing DialCoT-M and DialCoT-A,
the former opts for a sequential approach to answering sub-questions, 
as opposed to addressing them all-at-once. 
The superior performance metrics of DialCoT-M 
in comparison to DialCoT-A substantiate the claim that
a sequential methodology for answering sub-questions 
yields greater efficacy than an all-at-once approach.
(3) The primary difference between DialCoT-S and DialCoT-M 
is that DialCoT-S generates sub-questions sequentially
rather than generating all sub-questions at once.
Given DialCoT-S's stronger performance metrics, 
this indicates that 
the approach of sequentially decomposing sub-questions 
is more effective than generating all sub-questions in a single step.

\end{document}